 \documentclass[final,1p,times,authoryear]{elsarticle}

\usepackage{amssymb}
\usepackage{lipsum}

\usepackage{fullpage}
\usepackage{wrapfig}
\usepackage{comment}
\usepackage{amsmath}
\usepackage[utf8]{inputenc} 
\usepackage[T1]{fontenc}    
\usepackage{hyperref}       
\usepackage{url}            
\usepackage{booktabs}       
\usepackage{siunitx} 
\usepackage{amsfonts}       
\usepackage{nicefrac}       
\usepackage{microtype}      
\usepackage{xcolor}         
\definecolor{bestcolor}{rgb}{1,0,0} 
\usepackage{subcaption}
\usepackage{caption}
\usepackage[normalem]{ulem}
\usepackage{tikz}
\usepackage{comment}

\usepackage{amsfonts}       
\usepackage{amssymb}
\usepackage{amsmath}
\usepackage{amsthm}
\usepackage{mathtools}
\usepackage{multicol}
\usepackage{multirow}

\usepackage{color}

\usepackage{array}
\usepackage{algorithm}
\usepackage{algpseudocode}
\usepackage[shortlabels]{enumitem}

\usepackage{longtable}
\usepackage{booktabs}
\usepackage{geometry}
\usepackage{subcaption}
\usepackage{hyperref}
\graphicspath{{./Figures/} } 
\usepackage[normalem]{ulem}
\newcounter{ToDo}
\newcounter{gaocomm} 
\newcounter{Note}
\definecolor{blue-violet}{rgb}{0.00,0.75,0.90}
\definecolor{mygreen}{rgb}{0.0, 0.5, 0.0}
\definecolor{awesome}{rgb}{1.0, 0.13, 0.32}
\definecolor{bostonuniversityred}{rgb}{1.0, 0.0, 0.0}

\usepackage[symbol]{footmisc}

\newtheorem{lemma}{Lemma}
\newtheorem{proposition}{Proposition}
\newtheorem{corollary}{Corollary}
\theoremstyle{definition}

\journal{Not sure.}

\begin{document}

\begin{frontmatter}



\title{ST-MambaSync: The Complement of Mamba and Transformers for Spatial-Temporal in Traffic Flow Prediction}


\author[first]{Zhiqi Shao}
\affiliation[first]{organization={Business Analytics},
            addressline={University of Sydney}, 
            city={Sydney},
            postcode={2006}, 
            state={NSW},
            country={Australia}\\
            {zhiqi.shao; junbin.gao; ze.wang@sydney.edu.au}
}

\author[second]{Xusheng Yao}
\affiliation[second]{organization={College of Management and Economics},
            addressline={Tianjin University}, 
            city={Tianjin},
            postcode={300072}, 
            country={China}}
.
\author[third]{Ze Wang}
\affiliation[second]{organization={ITLS, },
addressline={University of Sydney}, 
            city={Sydney},
            postcode={2006}, 
            state={NSW},
            country={Australia}}

\author[first]{Junbin Gao}

\begin{abstract}

Accurate traffic flow prediction is pivotal for optimizing traffic management, enhancing road safety, and reducing environmental impacts. Traditional models, however, struggle with the demands of processing long sequence data, requiring substantial memory and computational resources, and suffer from slow inference times due to the lack of a unified summary state. This paper introduces ST-MambaSync, a novel traffic flow prediction model that integrates transformer technology with the innovative ST-Mamba block, marking a significant advancement in the field. We are the first to implement the Mamba mechanism—essentially an attention mechanism combined with ResNet—within a transformer framework, significantly improving the model's explainability and performance. ST-MambaSync not only addresses key challenges related to data length and computational efficiency but also sets new standards for accuracy and processing speed through extensive comparative analysis. This breakthrough has profound implications for urban planning and real-time traffic management, establishing a new benchmark in traffic flow prediction technology.

\end{abstract}




\end{frontmatter}




\section{Introduction}\label{Introduction}

Accurate traffic flow prediction is pivotal for optimizing traffic management, enhancing road safety, and reducing environmental impacts. Traditional models, however, struggle with the demands of processing long sequence data, requiring substantial memory and computational resources, and suffer from slow inference times due to the lack of a unified summary state. This paper introduces ST-MambaSync, a novel traffic flow prediction model that integrates transformer technology with the innovative ST-Mamba block, marking a significant advancement in the field. We are the first to implement the Mamba mechanism—essentially an attention mechanism combined with ResNet—within a transformer framework, significantly improving the model's explainability and performance. ST-MambaSync not only addresses key challenges related to data length and computational efficiency but also sets new standards for accuracy and processing speed through extensive comparative analysis. This breakthrough has profound implications for urban planning and real-time traffic management, establishing a new benchmark in traffic flow prediction technology.

\subsection{Traditional Approaches for Traffic Flow Predictions}

The literature on traffic flow prediction has historically relied on several traditional methodologies that have been instrumental in laying the foundational understanding of this field. Early approaches include the Historical Average (HA) method \citep{smith1995forecasting}, which predicts future traffic by averaging past data. While straightforward, this method often fails to capture sudden or atypical changes in traffic patterns.

Further developments led to the adoption of the Autoregressive Integrated Moving Average (ARIMA) model, introduced in traffic studies by \citep{kumar2015short}. ARIMA models enhance forecasting accuracy by considering past values and their errors; however, they inherently assume linear relationships among data points, which limits their effectiveness in handling the non-linear dynamics typical of traffic flows.

Another significant approach is the Support Vector Regression (SVR) model, utilized by \citep{wu2004travel} for traffic prediction. SVR has been preferred for its ability to handle non-linear data through the use of kernel functions. Despite its advantages, SVR, like its predecessors, struggles to manage the high-dimensional and complex relationships present in traffic datasets.
In addition to these methods, the k-Nearest Neighbors (k-NN) algorithm has also been applied to traffic flow prediction \citep{KNEAR_2020}. The k-NN method predicts traffic by considering the k most similar historical traffic patterns. While it can capture non-linear relationships, it is sensitive to the choice of k and may struggle with high-dimensional data.

Moreover, researchers have explored the use of Kalman Filtering (KF) for traffic prediction \citep{GUOKALMAN2014}. KF is a recursive algorithm that estimates the state of a system based on noisy measurements. It has been used to model the dynamic nature of traffic flows, but its performance can be affected by the quality of the initial estimates and the assumptions made about the system's behavior.

While these traditional models have provided valuable insights and served as a stepping stone in the evolution of traffic flow prediction, their simplicity and linear assumptions often fall short in complex and dynamic traffic scenarios. This limitation is primarily due to their inability to fully capture the nonlinear interactions and dependencies inherent in traffic data, thus potentially compromising the accuracy of the predictions in more challenging environments. As traffic systems become increasingly complex, the need for more sophisticated, non-linear models that can effectively handle these complexities becomes evident, paving the way for the next generation of traffic prediction methodologies.

\subsection{Deep Learning Method in Traffic Flow Prediction}
In recent years, the field of traffic flow prediction has seen notable progress with the integration of deep learning technologies. Deep learning models like Convolutional Neural Networks (CNNs) \citep{Sayed2023} and Recurrent Neural Networks (RNNs) have been effective in understanding spatial and temporal patterns. CNNs are adept at handling spatial data but often falter with long-range temporal patterns. Conversely, RNNs, including variants like Long Short-Term Memory (LSTM) and Gated Recurrent Units (GRU), are good at temporal dynamics but struggle with very long-term dependencies due to issues like the vanishing gradient problem, which impedes their performance on lengthy sequences. Moreover, the inherently sequential processing of RNNs restricts their parallelization, slowing down both training and inference compared to CNNs.

Transformers, originally designed for natural language processing, have recently been adapted for traffic flow prediction, introducing a new method for dependency recognition via self-attention mechanisms. These mechanisms allow Transformers to evaluate the significance of different segments of input data, enabling them to handle complex dependencies more effectively than RNNs and without the need for sequential data processing. This capacity to manage long-range dependencies has shown potential in enhancing the accuracy of predictions for long-term traffic flows.

Despite their advantages, Transformers also face significant challenges. The self-attention mechanism, while effective, requires substantial computational resources, especially when applied to large traffic networks and extensive historical data. The computational demands of Transformers increase quadratically with the length of the input sequence, which can slow down both training and testing. The high computational needs may also limit the practicality of Transformers in real-time traffic management systems that require fast predictions.

\subsection{State Space Model}
Given the limitations of existing deep learning methods, the Selective State of Space model (commonly referred to as Mamba) \citep{gu2023mamba} stands out for its ability to deliver high-accuracy forecasts while requiring less computational effort. This efficiency is particularly crucial in both short-term and long-term traffic management scenarios, where quick and reliable predictions are vital for effective congestion control, route optimization, and traffic regulation. A recent study \citep{shao2024stssms} marked the first to apply the Mamba model to spatial-temporal traffic flow prediction, demonstrating promising results in reducing computational costs. However, there remains room for improvement in balancing accuracy with manageable computational demands.

This paper introduces the Spatial-Temporal Mamba Transformer (ST-MambaSync), a novel framework that efficiently integrates the popular transformer and Mamba methods for accurate traffic flow prediction. The ST-MambaSync model comprises two main components: the ST-Transformer and the ST-Mamba Block. The ST-Transformer efficiently processes data, capturing global information through spatial and temporal features. In contrast, the ST-Mamba Block, which includes an ST-Mixer, converts the tensor into a matrix. This matrix is then fed into the ST-Mamba layer, which updates individual hidden states and extends memory for long-range data, focusing more on local information. The combination on ST-Transformer and ST-Mamba Block not only enhances both global and local features but also accelerates computation, making it an effective component of our integrated approach for managing complex traffic data.

\subsection{Contribution}
To the best of our knowledge, this paper introduces a groundbreaking integration of the selective state-of-space model (Mamba) with attention mechanisms specifically tailored for spatial-temporal data, establishing that Mamba effectively functions as a type of attention within a ResNet framework. The principal contributions of our research are summarized as follows:

\begin{itemize}
\item This study is the first to combine Mamba and attention blocks to manage spatial-temporal data, enhancing both real-time and long-term traffic forecasting.
\item We provide theoretical evidence showing that the Mamba model operates as an attention mechanism within a ResNet framework. This integration with the Transformer and the ST-Mamba block significantly boosts the model’s capacity to assimilate both comprehensive and granular information, thereby improving its performance in handling complex datasets.
\item Through rigorous testing on real-world traffic datasets, our model demonstrates superior performance relative to existing benchmarks, achieving notable gains in accuracy and efficiency while lowering computational demands.
\end{itemize}

This innovative approach not only advances the field of traffic flow prediction but also sets a new standard for the application of hybrid models in complex data environments.

\section{Preliminary and Problem Statement}\label{Prelimanary}
\subsection{Notations}
\paragraph{Road Network} In this study, we define a road network as a graph $\mathcal{G} = (\mathcal{V}, \mathcal{E})$, where the node set $\mathcal{V} = {v_1, \ldots, v_N}$ corresponds to $N$ critical points such as traffic sensors and intersections on the roads. Each element of the edge set $\mathcal{E} \subseteq \mathcal{V} \times \mathcal{V}$ represents a section of road. Within this network, $\mathbf{X}_t \in \mathbb{R}^{N \times d}$ represents the traffic flow at the $N$ nodes at any given timestamp $t$, encapsulating flow characteristics across these nodes, with $d$ representing the dimensions of the features. Over a period $T>0$, the traffic data is aggregated into a mode-3 tensor $\mathbf{X} \in \mathbb{R}^{T \times N \times d}$, organizing the data along temporal, spatial, and feature-specific dimensions.

\paragraph{Problem Statement} The objective of traffic flow forecasting is to predict future traffic conditions accurately using historical data. For this purpose, we define a function $f$ within the context of a road network $\mathcal{G}$, which leverages traffic flow data from the past $M$ timestamps to forecast traffic conditions over the next $Z$ timestamps. This relationship can be mathematically formulated as:
\begin{equation*}
f\left( [ \mathbf{X}_{t-M+1}, \ldots, \mathbf{X}_{t}] ; \mathcal{G} \right) \mapsto [ \mathbf{X}_{t+1}, \ldots, \mathbf{X}_{t+Z}],
\end{equation*}
where $\mathbf{X}_{t}$ denotes the traffic flow tensor at time $t$. Throughout the model learning phase, we define $t$ to range from $M$ to $T-Z$ in order to efficiently utilize the observed traffic flow tensor $\mathbf{X}$.

\subsection{Attention}
\label{Attention}
The self-attention mechanism, originally conceived for natural language processing (NLP), enriches the representation of feature data by revealing the underlying ``self-attention'' within the data set. Beginning with the input \( X\in \mathbb{R}^{N\times d} \), the mechanism constructs queries (\( Q \)), keys (\( K \)), and values (\( V \)) via transformation matrices:
\begin{equation}
Q = X W_Q, \quad K = X W_K, \quad V = X W_V,
\end{equation}
where \( W_Q \in \mathbb{R}^{d\times d_q} \), \( W_K \in \mathbb{R}^{d\times d_k} \), \( W_V \in \mathbb{R}^{d\times d_v} \) are weight matrices that are learned, and \( d_q=d_k=d_v=d_0 \) for simplicity.

Attention scores are computed by the scaled dot-product of queries and keys:
\begin{equation}
A = \frac{{Q}{K}^\top}{\sqrt{d_k}},
\end{equation}
where scaling by \( \sqrt{d_k} \) provides numerical stability.

The attention scores \( A \) are normalized using the softmax function to obtain the attention weights \( O \):
\begin{equation}
O = \text{softmax}(A).
\end{equation}

The final representation \( Y \) emerges as the weighted sum of the values:
\begin{equation}
 Y = O V.
\end{equation}

Multi-head attention leverages multiple ``heads'' of \( Q \), \( K \), and \( V \) to explore different representation subspaces, creating a rich, integrated output.

\textit{Remark 1:} We standardize \( d_q=d_k=d_v=d_0 \) to ensure uniform dimensions across the attention mechanism's architecture.

\subsection{State of Space Model}\label{SSMS_PRE}

The state-space representation of a continuous-time linear time-invariant system can be described by the following differential equations:
\begin{align}
    h(t) &= Ah(t-1) + Bu(t-1) \\
    y(t) &= Ch(t) + Du(t)
\end{align}

The solution to the state equation over a time interval can be given by:
\begin{align}
    h(t_b) &= e^{A(t_b - t_a)}h(t_a) + \int_{t_a}^{t_b} e^{A(t_b - \tau)}B(\tau)u(\tau) \, d\tau
\end{align}

In discrete time-steps, the state at step \( b \) is given by:
\begin{align}
    h_b &= e^{A(\Delta_a + \dots + \Delta_{b-1})} \left( h_a + \sum_{i=a}^{b-1}B_iu_ie^{-A(\Delta_a + \dots + \Delta_i)}\Delta_i \right)
\end{align}

For the transition from state \( h_a \) to state \( h_{a+1} \), the following discrete update can be used:
\begin{align}\label{diss_ssms}
    h_{a+1} &= e^{A\Delta_a} \left( h_a + B_a u_a e^{-A\Delta_a} \right) \\
             &= e^{A\Delta_a}h_a + B_a u_a \\
             &= \hat{A}_ah_a + \hat{B}_au_a
\end{align}

\section{Method}\label{Method}

\begin{figure}[h]
    \centering
 \includegraphics[scale=0.3]{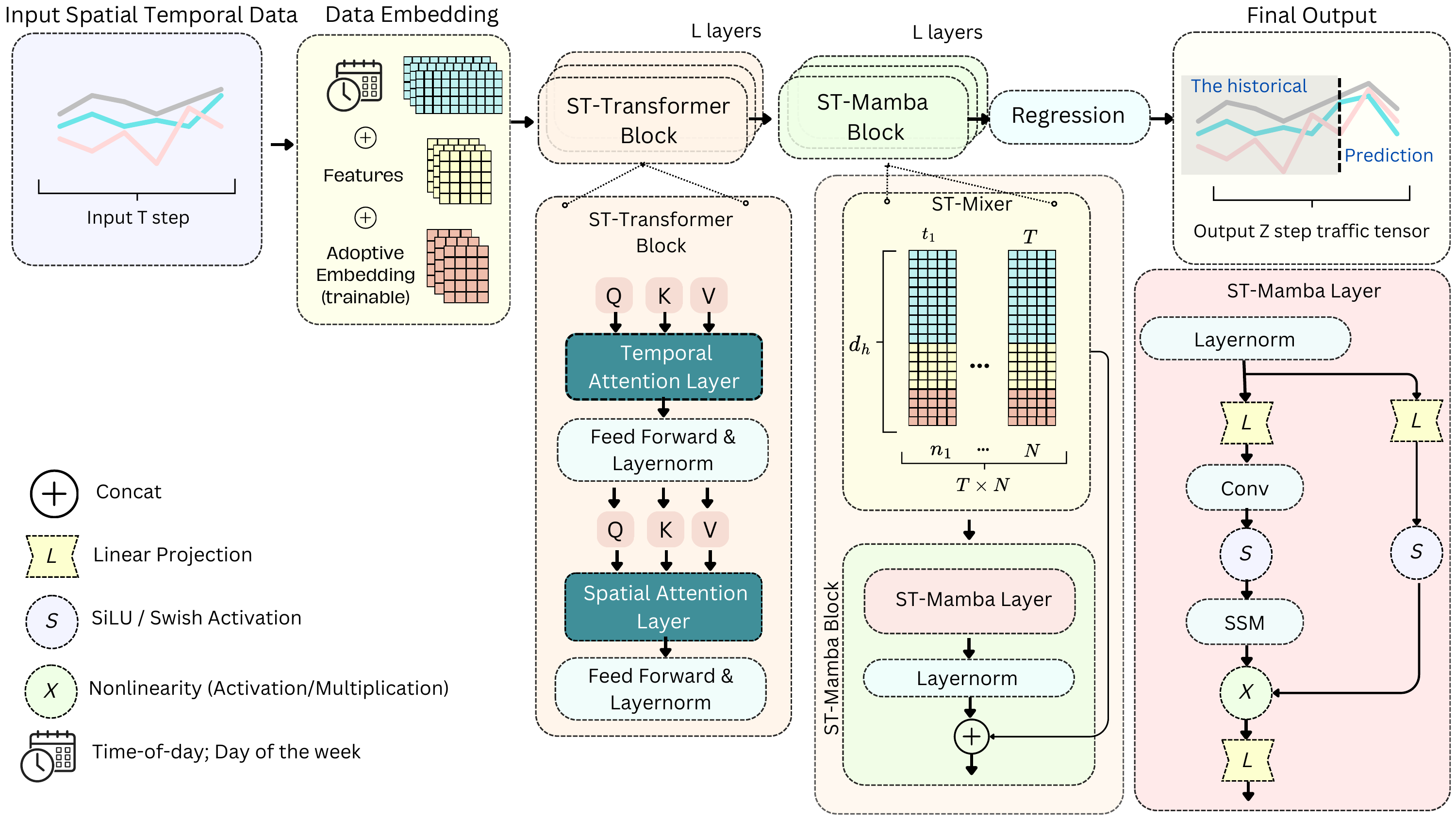}
    \caption{The framework of proposed ST-MambaSync.}
    \label{fig:framework}
\end{figure}

\subsection{Data Embedding}\label{Embedding}
To encapsulate and reflect the dynamic temporal patterns present within the traffic data, we utilized a modifiable data embedding layer that processes the sequential input $\mathbf{X}_{t-M+1:t}$. Through the application of a dense neural layer, we extract the intrinsic feature embedding $\mathbf{Z}^f_t \in \mathbb{R}^{M \times N \times d_e}$:

\begin{equation}
\mathbf{Z}^f_t = \text{Dense}(\mathbf{X}_{t-M+1:t})
\end{equation}

Here, $d_e$ signifies the dimension of the embedded features, and $\text{Dense}(\cdot)$ represents the applied dense layer. Furthermore, we introduce a parameterized dictionary for the embedding of weekdays $\mathbf{Z}_w \in \mathbb{R}^{7 \times d_e}$ and another for the embedding of distinct times of the day $\mathbf{Z}_h \in \mathbb{R}^{288 \times d_e}$, encapsulating the cyclical nature of weeks with 7 days and days with 288 time intervals. With $\mathbf{W}_t \in \mathbb{R}^{M}$ representing the weekday index and $\mathbf{H}_t \in \mathbb{R}^{M}$ representing the time-of-day index over the period from $t-M+1$ to $t$, we map these indices to their respective embeddings, yielding the weekday embedded features $\mathbf{Z}_{w_t} \in \mathbb{R}^{M \times d_e}$ and time-of-day embedded features $\mathbf{Z}_{h_t} \in \mathbb{R}^{M \times d_e}$. The combination and expansion of these embeddings generate the cyclical feature embedding $\mathbf{Z}^c_t \in \mathbb{R}^{M \times N \times 2d_e}$, which is utilized to incorporate periodic patterns into the traffic data.

Considering the rhythmic progression of time and the interlinked nature of traffic events, traffic sensors yield data with unique temporal traits. To address the need for a uniform approach to encapsulate these spatio-temporal dynamics, a shared spatio-temporal adaptive embedding, $\mathbf{Z}^s_t \in \mathbb{R}^{M \times N \times d_s}$, is put forth. This embedding is initialized utilizing Xavier uniform initialization, a technique that primes the model's weights to avoid excessively large or small gradients initially, and thereafter, it is treated as a model parameter.

The integration of the aforesaid embeddings results in a hidden spatio-temporal representation $\mathbf{Z} \in \mathbb{R}^{M \times N \times d_h}$:

\begin{equation}
\mathbf{Z}_t = \textbf{Concatenate}(\mathbf{Z}^f
_t; \mathbf{Z}^c_t ;\mathbf{Z}^s_t)
\end{equation}

In this equation, the concatenation operation is denoted by a comma, and the dimension of the hidden representation $d_h$ is computed as $3d_e + d_s$.

\subsection{Spatial Temporal Transformer (ST-Transformer Block)}\label{ST-Transformer Block}
We utilize standard transformers along both temporal and spatial dimensions to understand complex traffic interactions. Given a hidden spatio-temporal matrix $\mathbf{Z} \in \mathbb{R}^{T \times N \times d_h}$, where $T$ is the number of frames and $N$ represents spatial nodes, we derive the query, key, and value matrices using temporal transformer layers as follows:
\begin{align}
\mathbf{Q}^{(te)} &= \mathbf{Z} \mathbf{W}_Q^{(te)}, \\
\mathbf{K}^{(te)} &= \mathbf{Z} \mathbf{W}_K^{(te)}, \\
\mathbf{V}^{(te)} &= \mathbf{Z} \mathbf{W}_V^{(te)},
\end{align}
where $\mathbf{W}_Q^{(te)}, \mathbf{W}_K^{(te)}, \mathbf{W}_V^{(te)} \in \mathbb{R}^{d_h \times d_h}$ are learnable parameters. The self-attention scores are computed as:
\begin{equation}
\mathbf{A}^{(te)} = \text{Softmax}\left(\frac{\mathbf{Q}^{(te)} (\mathbf{K}^{(te)})^\top}{\sqrt{d_h}}\right),
\end{equation}
capturing temporal connections across different spatial nodes. The output of the temporal transformer, $\mathbf{Z}^{(te)} \in \mathbb{R}^{T \times N \times d_h}$, is then obtained as:
\begin{equation}
\mathbf{Z}^{(te)} = \mathbf{A}^{(te)} \mathbf{V}^{(te)}.
\end{equation}
In a similar fashion, the spatial transformer layer functions by processing $\mathbf{Z}^{(te)}$ through self-attention (following the same equations) to produce $\mathbf{Z}^{(sp)} \in \mathbb{R}^{T \times N \times d_h}$. Important enhancements include layer normalization, residual connections, and a multi-head mechanism.

\subsection{Spatial Temporal Selective State of Spatial (ST-Mamba block)}\label{ST-Mamba}
As depicted in Figure~\ref{fig:framework}, following the adaptive ST-Transformer block, our framework employs a simplified ST-Mamba block. This block features an ST-Mamba layer designed to reduce computational costs and to enhance long-term memory. To feed the input to ST-Mamba block, we employ a tersor reshape named as ST-mixer as: 
\paragraph{ST-mixer}
To effectively blend spatial and temporal data, the ST-SSMs utilize tensor reshaping, as detailed in Figure~\ref{fig:framework}, to transform tensor \( \mathbf{Z}^{(sp)} \) into matrix \( \mathbf{\bar{X}} \). This transformation involves aligning and concatenating the tensor slices across each time step \( t \) as follows:
\begin{align}
    \mathbf{\bar{X}} = \text{reshape}(\mathbf{Z}^{(sp)}).
\end{align}
Through this reshaping, we obtain a new embedding \( \mathbf{\bar{X}} \) in \( \mathbb{R}^{\mathcal{T} \times d_h} \), where \( \mathcal{T} \) encapsulates the total temporal length \( T \times N \), representing the spatial dimension. This reshaping facilitates the unified processing of spatial and temporal information, thereby capturing complex patterns in the data more effectively.

\subsubsection{ST-Mamba Layer}
The ST-Mamba layer, as described in Section~\ref{SSMS_PRE}, utilizes the discretization of continuous state-space models (SSM). We denote the input to the ST-Mamba block as $\mathcal{H}$, which is obtained by applying LayerNorm to $\mathbf{\bar{X}}$. This processed input is then subjected to the selective state-space model (SSM) layer, where a linear transformation produces:
\begin{equation}
    \mathbf{U} = Linear(\mathcal{H}),
\end{equation}
with $\mathbf{U} \in \mathbb{R}^{d_h \times \mathcal{T}}$ representing the hidden state's latent representation at each iteration step $k$. The objective is to calculate the output $\mathbf{Y} \in \mathbb{R}^{d_{inner} \times \mathcal{T}}$, which is then projected back to match the input's dimensions $\mathcal{T} \times d_h$.

\paragraph{Parameter Initialization}
Initializing parameters in the ST-SSM is vital for its operation:
\begin{itemize}
    \item $\mathbf{A} \in \mathbb{R}^{d_{inner} \times d_{state}}$: This structured state transition matrix is initialized using HiPPO to ensure capturing of long-range dependencies.
    \item $\mathbf{B} \in \mathbb{R}^{d_{state} \times \mathcal{T}}$: Calculated as $\mathbf{B} = s_B(\mathbf{U})$, where $s_B(\cdot)$ is a learnable linear projection.
    \item $\mathbf{C} \in \mathbb{R}^{d_{state} \times \mathcal{T}}$: Output projection matrix, derived as $\mathbf{C} = s_C(\mathbf{U})$, where $s_C(\cdot)$ is a learnable projection.
    \item $\mathbf{D} \in \mathbb{R}^{d_{inner}}$: A learnable parameter that facilitates the direct transfer of information from input to output, bypassing the state transformation.
    \item $\Delta \in \mathbb{R}^{d_{state} \times \mathcal{T}}$: Step size parameter, determined using $\Delta = \tau_\Delta(\text{Parameter} + s_\Delta(\mathbf{U}))$, with $\tau_\Delta$ being the softplus function and $s_\Delta(\cdot)$ a linear projection.
\end{itemize}

\paragraph{Discretization and Output Computation}
The conversion of continuous-time parameters into discrete-time SSM parameters involves:
\begin{align}
    \mathbf{\tilde{A}} &= \exp(\Delta \mathbf{A}), \\
    \mathbf{\tilde{B}} &= \mathbf{A}^{-1} (\exp(\Delta \mathbf{A}) - \mathbf{I}) \mathbf{B},
\end{align}
where $\exp(\cdot)$ is the matrix exponential function, and $\mathbf{I}$ represents the identity matrix of suitable size. These discrete-time matrices, $\mathbf{\tilde{A}}$ and $\mathbf{\tilde{B}}$, facilitate the recurrence within the selective ST-Mamba layer:
\begin{align}
    \mathbf{H}_k &= \mathbf{\tilde{A}} \odot \mathbf{H}_{k-1} + \mathbf{\tilde{B}} \odot \mathbf{U}_k, \\
    \mathbf{Y}_k &= \mathbf{C} \odot \mathbf{H}_k + \mathbf{D} \odot \mathbf{U}_k,
\end{align}
with $\odot$ denoting the Hadamard product. This iterative process spans each step from $k = 1$ to $\mathcal{T}$, ensuring that each data step is transformed through the SSM layer. The final output, $\mathbf{Y}$, is reshaped into $\mathbb{R}^{\mathcal{T} \times d_{inner}}$ to align with the original input dimensions.

\subsubsection{ST-Mamba block and Regression Layer}

\paragraph{Normalization Layer}
Within the ST-Mamba block, the normalization of layers is crucial for improving the training's stability and efficiency. Specifically, consider an input matrix \(\mathbf{Y}\) with dimensions \(\mathbb{R}^{\mathcal{T} \times d_h}\), where \(\mathcal{T}\) represents the sequence length or sample count, and \(d_h\) indicates the features' dimensional space. The normalization process is described by:
\begin{align}
    \text{Normalization}(\mathbf{Y}) = \gamma \odot \frac{\mathbf{Y} - \mu}{\sqrt{\sigma^2 + \epsilon}} + \beta.
\end{align}
In this formula, \(\mu\) and \(\sigma^2\) are the mean and variance computed along the features' dimension \(d_h\), resulting in vectors of size \(\mathcal{T} \times 1\). The scale (\(\gamma\)) and shift (\(\beta\)) parameters, each sized \(1 \times d_h\), are adjustable, optimizing the normalization's impact. This process ensures stability in the model's learning phase while allowing the reintegration of the original activations distribution if it improves model performance. The addition of \(\epsilon\), a small constant, prevents any division by zero, maintaining numerical stability. Through layer normalization, the model effectively reduces internal covariate shift, enhancing training speed and boosting overall deep learning performance.

\paragraph{Regression Layer}
As in Referring to Figure~\ref{fig:framework}, the output from the ST-Mamba block passes through a normalization step before reaching the regression layer, which is structured as follows:
\begin{align}
    \mathbf{\bar Y}  & = \text{Normalization}(\mathbf{Y}) +  \mathbf{\bar X}\\
    \mathcal{Y} & = FC(\mathbf{\bar Y})
\end{align}
In this configuration, \(FC(\cdot)\) denotes the fully connected layer that processes the normalized data. The resultant \(\mathcal{Y}\), existing within the dimensional space \(\mathbb{R}^{Z \times N \times d}\), marks the culmination of the process. This structured approach showcases how architectural enhancements are designed to improve the training of deep networks and accurately interpret complex, multi-dimensional datasets.

\subsection{Analysis of Mamba and Attention}
\begin{lemma}[Analogy Between Attention and Linear Regression]
Given a dataset $\{(x_i, y_i)\}_{i=1}^N$ where $x_i \in \mathbb{R}^d$ and $y_i \in \mathbb{R}$, consider a linear regression model defined by $y = x_i^T \omega$, where $\omega \in \mathbb{R}^d$. The least squares solution for the model coefficients $\hat{\omega}$ is expressed as $\hat{\omega} = (X^TX)^{-1}X^T\mathbf{y}$, enabling predictions $\hat{y}$ on the training data to be given by $\hat{y} = X\hat{\omega} = X(X^TX)^{-1}X^T\mathbf{y}$. This form of linear prediction can be interpreted as an attention mechanism, where each prediction is a weighted sum of all outputs, with weights determined by the matrix multiplication $X(X^TX)^{-1}X^T$.
\end{lemma}

\begin{proof}
For the given data $\{(x_i, y_i)\}_{i=1}^N$, define the matrix $X$ and vector $\mathbf{y}$ as:
\begin{equation}
    X = \begin{bmatrix}
    x_1^T \\
    \vdots \\
    x_N^T
    \end{bmatrix} \in \mathbb{R}^{N \times d}, \quad
    \mathbf{y} = \begin{bmatrix}
    y_1 \\
    \vdots \\
    y_N
    \end{bmatrix} \in \mathbb{R}^N.
\end{equation}

The least squares solution is given by:
\begin{equation}
    \hat{\omega} = (X^TX)^{-1}X^T\mathbf{y}.
\end{equation}
Using this solution, the model prediction on the training data is:
\begin{equation}
    \hat{y} = X\hat{\omega} = X(X^TX)^{-1}X^T\mathbf{y},
\end{equation}
where each element $\hat{y}_i$ for $i = 1, \ldots, N$ can be expressed as:
\begin{equation}
    \hat{y}_i = \sum_{j=1}^N a_{ij} y_j,
\end{equation}
with $a_{ij} = x_i^T (X^TX)^{-1}x_j$. These coefficients $a_{ij}$ are analogous to attention scores, typically denoted as $QK^T$ in attention mechanisms. Each $\hat{y}_i$ represents a weighted average over all outputs, where the weights are the similarities (or 'attention') the model pays to each training value based on $x_i$.

For a new data point $x'$, the prediction $\hat{y}(x')$ by the model is:
\begin{equation}
    \hat{y}(x') = x' (X^TX)^{-1}X^T\mathbf{y},
\end{equation}
which can be interpreted as the weighted sum of the training data targets, where the weights are the similarities between the new data point $x'$ and the training data points, akin to $QK^T$, with $\mathbf{y}$ analogous to $V$ in attention mechanisms.

\end{proof}

\begin{proposition}[Extension to Spatial State Models]
In the context of Spatial State Models (SSMs), the discrete update equation can be modeled through dynamics analogous to attention mechanisms. Specifically, for each state update:
\[
h_{a+1} = e^{A\Delta_a} h_a + B_a u_a e^{-A\Delta_a} \Delta A e^{A\Delta_a},
\]
where $Q_m = B_a$, $K_m = u_a$, and $W = e^{A\Delta_a}$ play roles analogous to query, key, and weight transformations in attention mechanisms. This results in an attention-like score for each input that directly influences the model's state updates.
\end{proposition}

\begin{proof}
For each discrete update, the SSMs are actually finding an attention score for each of the inputs $u_a$ and a learnable linear projection of the input $u_a$ (denoted as $B_a$).

To find the final results, we define \( V_m \) as \( C_a \), representing the values in the attention mechanism, and then:
\begin{equation}\label{eq:att
_mamba}
    y = W^{-1}\left(Q_mK_m^T\right) V_m+ WV_m\beta h_{a+1}.
\end{equation}
The first term resembles the attention mechanism, capturing the weighted importance of different inputs. The second term incorporates the updated hidden state, structured analogously to a Residual Network (ResNet), where the hidden state is updated by adding a transformed version of itself.
\end{proof}

\begin{corollary}
The attention mechanism formalism enables SSMs to perform weighted sums of inputs, akin to how attention in neural networks aggregates information. This approach enhances the ability of SSMs to dynamically adjust to changing inputs and efficiently compute new states, facilitating more robust predictions and state estimations.
\end{corollary}

\subsection{The iteration expression on ST-MambaSync}

In ST-MambaSync, the input vector $\mathbf{Z}$ initially passes through an attention layer and then proceeds to an ST-Mamba Layer. The transformer component primarily captures global information, while the integration with the ST-Mamba block, proven to be an attention mechanism combined with ResNet, complements this by focusing on local details. This combination effectively enhances the transformer's ability to manage both global and local information, significantly improving the model's overall predictive accuracy and efficiency. 

\section{Experiment} \label{Experiment}
\subsection{Data Description and Baseline Models}
\paragraph{Datasets} We 
test our method 
on six major traffic forecasting benchmarks, namely METR-LA, PEMS-BAY, PEMS03, PEMS04, PEMS07, and PEMS08, to verify its effectiveness. The datasets under consideration feature a time resolution of 5 minutes, which results in 12 data frames being recorded for every hour. The detail of the data is summarized in Table~\ref{data}. 

\begin{table}[ht]
\centering
\caption{Summary of Datasets.}
\begin{tabular}{lccc}
\hline
Dataset    & \#Sensors (N) & \#Timesteps & Time Range         \\ \hline
METR-LA    & 207           & 34,272      & 03/2012 - 06/2012  \\
PEMS-BAY   & 325           & 52,116      & 01/2017 - 05/2017  \\
PEMS03     & 358           & 26,209      & 05/2012 - 07/2012  \\
PEMS04     & 307           & 16,992      & 01/2018 - 02/2018  \\
PEMS07     & 883           & 28,224      & 05/2017 - 08/2017  \\
PEMS08     & 170           & 17,856      & 07/2016 - 08/2016  \\ \hline
\end{tabular}\label{data}
\end{table}

\paragraph{Baseline Models} 

In our comparative analysis, we evaluate the performance of our proposed approach against a comprehensive set of baselines within the traffic forecasting domain. 
\begin{itemize}
    \item Historical Index (HI) \citep{cui2021historical}: serving as the conventional benchmark, reflecting standard industry practices.
\end{itemize}
 Our examination extends to a series of Spatial-Temporal Graph Neural Networks (STGNNs)—including
 \begin{itemize}
     \item GWNet \citep{wu2020connecting}: proposing a graph neural network framework that automatically extracts uni-directed relations among variables, addressing the limitation of existing methods in fully exploiting latent spatial dependencies in multivariate time series forecasting.
     \item DCRNN \citep{li2017diffusion}: introducing the Diffusion Convolutional Recurrent Neural Network for traffic forecasting which captures both spatial and temporal dependencies.
     \item AGCRN \citep{bai2020adaptive}: introducing adaptive modules to capture node-specific patterns and infer inter-dependencies among traffic series which provides fine-grained modeling of spatial and temporal dynamics in traffic data.
     \item STGCN \citep{ijcai2018p505}: proposing a deep learning framework that integrates graph convolutions for spatial feature extraction and gated temporal convolutions for temporal feature extraction.
     \item GTS \citep{shang2021discrete}: proposing a method for forecasting multiple interrelated time series by learning a graph structure simultaneously with a Graph Neural Network (GNN) which addresses the limitations of a previous method.
     \item MTGNN \citep{Wu2020ConnectingTD}: proposing a graph neural network framework that automatically extracts uni-directed relations among variables which captures both spatial and temporal dependencies.
    \item GMAN \citep{Zheng_Fan_Wang_Qi_2020}: introducing a graph-based deep learning model that incorporates spatial and temporal attention mechanisms to capture dynamic correlations among traffic sensors.
 \end{itemize}  
Recognizing the potential of Transformer-based models in time series forecasting, we particularly focus on:
\begin{itemize}

    \item PDFormer \citep{pdformer}: introducing a traffic flow prediction model that captures dynamic spatial dependencies, long-range spatial dependencies, and the time delay in traffic condition propagation.
    \item STAEformer \citep{staeformer}: proposing a spatio-temporal adaptive embedding that enhances the performance of vanilla transformers for traffic forecasting.
\end{itemize}  and which are adept at short-term traffic forecasting tasks. 
Additionally, we explore 
\begin{itemize}
    \item STNorm \citep{dengstnorm}: leveraging spatial and temporal normalization modules to refine the high-frequency and local components underlying the raw data.
    \item STID \citep{shaozhao2022}: proposing an approach that addresses the indistinguishability of samples in both spatial and temporal dimensions by attaching spatial and temporal identity information to the input data.
\end{itemize}
Those diverse range of models allows for a robust validation of our proposed method's capabilities.

\subsection{Experiment Setup}
\label{Experiment Setup}
\paragraph{Implementation} All experiments were carried out on a machine equipped with an RTX 3090 GPU (24GB) and a 15-core CPU. The data from the PEMS-BAY, PEMS03, PEMS04, PEMS07, and PEMS08 datasets were divided into training, validation, and test sets. PEMS-BAY was split in a 7:1:2 ratio, whereas PEMS03, PEMS04, PEMS07, and PEMS08 were divided using a 6:2:2 ratio. The embedding dimension (\(d_f\)) was set to 24 and the attention dimension (\(d_a\)) to 80. The model architecture includes a single layer for both spatial and temporal transformers, with four heads, and one ST-Mamba layer with the expansion dimension set to 2. Both the input and forecast horizon were defined as 1 hour, equivalent to \({M} = Z = 12\). Optimization was performed using the Adam optimizer, starting with a learning rate of 0.001 that gradually decreased, and a batch size of 16. To improve training efficiency, an early stopping mechanism was implemented, ceasing training if the validation error did not improve after 30 consecutive iterations.

\paragraph{Metric}
To evaluate the performance of traffic forecasting methods, three prevalent metrics are employed: the Mean Absolute Error (MAE), the Mean Absolute Percentage Error (MAPE), and the Root Mean Square Error (RMSE). These metrics offer a comprehensive view of model accuracy and error magnitude. They are defined as follows:
\begin{itemize}
    \item MAE (Mean Absolute Error): quantifies the average magnitude of the errors in a set of predictions, without considering their direction. It's calculated as:
  \begin{equation*}
    \text{MAE} = \frac{1}{n} \sum_{i=1}^{n} |\hat{y}_i - y_i|,
  \end{equation*}
  \item MAPE (Mean Absolute Percentage Error): expresses the error as a percentage of the actual values, providing a normalization of errors that is useful for comparisons across datasets of varying scales. It's given by:
  \begin{equation*}
    \text{MAPE} = \frac{1}{n} \sum_{i=1}^{n} \left| \frac{\hat{y}_i - y_i}{y_i} \right| \times 100,
  \end{equation*}
  \item RMSE (Root Mean Square Error): measures the square root of the average squared differences between the predicted and actual values, offering a high penalty for large errors. This metric is defined as:
  \begin{equation*}
    \text{RMSE} = \sqrt{\frac{1}{n} \sum_{i=1}^{n} (\hat{y}_i - y_i)^2}.
  \end{equation*}
\end{itemize}

In these equations, \(y = \{y_1, y_2, \ldots, y_n\}\) represents the set of ground-truth values, while \(\hat{y} = \{\hat{y}_1, \hat{y}_2, \ldots, \hat{y}_{n}\}\) denotes the corresponding set of predicted values. Through the utilization of MAE, MAPE, and RMSE, a thorough evaluation of model performance in forecasting traffic conditions can be achieved, highlighting not just the average errors but also providing insights into the distribution and proportionality of these errors relative to true values.

\subsection{Performance Evaluation}
To assess the effectiveness of the ST-MambaSync model, we utilized six real-world datasets varying significantly in complexity and scale. The datasets range from METR-LA, which includes 207 sensors, to PEMS07, encompassing 883 sensors. This selection provides a broad spectrum of urban traffic patterns and sensor network densities, potentially impacting the predictive performance of the model. The most outstanding results across these evaluations are denoted in \textcolor{red}{\textbf{red}} to highlight superior performance.

\paragraph{Analysis of Performance Metrics} The results are shown in Table~\ref{tab:performance_comparison_whole}, the ST-MambaSync model showcases its performance across four different PeMS datasets with a focus on three key metrics: MAE, RMSE, and MAPE. Particularly notable is its performance on the PEMS08 dataset, where ST-MambaSync achieves the lowest MAE of 13.30 and RMSE of 23.14, along with a MAPE of 8.80\%, indicating superior accuracy compared to other models. Although the model performs consistently across other datasets, such as PEMS03 and PEMS04 with competitive MAEs and RMSEs, it stands out in PEMS08, suggesting its effectiveness in environments with similar traffic patterns and sensor configurations.

\begin{table}[t]
\centering
\caption{Performance comparison of models on PEMS datasets, here we denote N as the number of sensors for each dataset.}
\label{tab:performance_comparison_whole}
\setlength{\tabcolsep}{3pt}
\begin{tabular}{l S[table-format=2.2] S[table-format=2.2] c S[table-format=2.2] S[table-format=2.2] c S[table-format=2.2] S[table-format=2.2] c S[table-format=2.2] S[table-format=2.2] c}
\toprule
{Model} & \multicolumn{3}{c}{PEMS03 (N=358)} & \multicolumn{3}{c}{PEMS04 (N=307)} & \multicolumn{3}{c}{PEMS07(N=883)} & \multicolumn{3}{c}{PEMS08(N=170)} \\
\cmidrule(lr){2-4} \cmidrule(lr){5-7} \cmidrule(lr){8-10} \cmidrule(lr){11-13}
& {MAE} & {RMSE} & {MAPE} & {MAE} & {RMSE} & {MAPE} & {MAE} & {RMSE} & {MAPE} & {MAE} & {RMSE} & {MAPE} \\
\midrule
HI & 32.62 & 49.89 & 30.60\% & 42.35 & 61.66 & 29.92\% & 49.03 & 71.18 & 22.75\% & 36.66 & 50.45 & 21.63\% \\
GWNet & \textcolor{red}{\textbf{14.59 }}& \textcolor{red}{\textbf{25.24}} & 15.52\% & 18.53 & 29.92 & 12.89\% & 20.47 & 33.47 & 8.61\% & 14.40 & 23.39 & 9.21\% \\
DCRNN & 15.54 & 27.18 & 15.62\% & 19.63 & 31.26 & 13.59\% & 21.16 & 34.14 & 9.02\% & 15.22 & 24.17 & 10.21\% \\
AGCRN & 15.24 & 26.65 & 15.89\% & 19.38 & 31.25 & 13.40\% & 20.57 & 34.40 & 8.74\% & 15.32 & 24.41 & 10.03\% \\
STGCN & 15.83 & 27.51 & 16.13\% & 19.57 & 31.38 & 13.44\% & 21.74 & 35.27 & 9.24\% & 16.08 & 25.39 & 10.60\% \\
GTS & 15.41 & 26.15 & 15.39\% & 20.96 & 32.95 & 14.66\% & 22.15 & 35.10 & 9.38\% & 16.49 & 26.08 & 10.54\% \\
MTGNN & 14.85 & 25.23 & 14.55\% & 19.17 & 31.70 & 13.37\% & 20.89 & 34.06 & 9.00\% & 15.18 & 24.24 & 10.20\% \\
STNorm & 15.32 & 25.93 & 14.37\% & 18.96 & 30.98 & 12.69\% & 20.50 & 34.66 & 8.75\% & 15.41 & 24.77 & 9.76\% \\
GMAN & 16.87 & 27.92 & 18.23\% & 19.14 & 31.60 & 13.19\% & 20.97 & 34.10 & 9.05\% & 15.31 & 24.92 & 10.13\% \\
PDFormer & 14.94 & 25.39 & 15.82\% & 18.36 & 30.03 & 12.00\% & 19.97 & 32.95 & 8.55\% & 13.58 & 23.41 & 9.05\% \\
STID & 15.33 & 27.40 & 16.40\% & 18.38 & 29.95 & 12.04\% & 19.61 & 32.79 & 8.30\% & 14.21 & 23.28 & 9.27\% \\
STAEformer& 15.35 & 27.55 & \textcolor{red}{\textbf{15.18\%}}& 18.22 & 30.18 & \textcolor{red}{\textbf{11.98\%}} & \textcolor{red}{\textbf{19.14}} & 32.60 & 8.01\% & 13.46 & 23.25 & 8.88\% \\
\hline
ST-MambaSync & 15.30 & 27.47 & \textcolor{red}{\textbf{15.18\%}}  & \textcolor{red}{\textbf{18.20
}} & \textcolor{red}{\textbf{29.85}} & 12.00\% & \textcolor{red}{\textbf{19.14}}  & \textcolor{red}{\textbf{32.58}} & \textcolor{red}{\textbf{7.97\%}}   &\textcolor{red}{\textbf{13.30}}	& \textcolor{red}{\textbf{23.14}}	& \textcolor{red}{\textbf{8.80 \%}}\\
\bottomrule
\end{tabular}
\end{table}

\paragraph{Detailed Analysis Based on Time Horizons} As the performance results showing in Table~\ref{tab:horizon}, the models are evaluated across multiple time horizons (15 minutes, 30 minutes, 60 minutes) on the METR-LA and PEMS-BAY datasets. the ST-MambaSync model demonstrates competitive accuracy, particularly excelling in the PEMS-BAY dataset across all time horizons. The visual comparison helps to underscore differences in performance stability and prediction accuracy between models, reaffirming the strengths of ST-MambaSync in shorter forecasting intervals.

\begin{table}[t]
\footnotesize
\centering
\caption{Performance on METR-LA and PEMS-BAY, we denote N as the number of sensors for each dataset.}
\setlength{\tabcolsep}{1.2pt} 
\begin{tabular}{llcccccccccccccc}
\hline
Horizon & Metric & HI & GWNet & DCRNN & AGCRN & STGCN & GTS & MTGNN & STNorm & GMAN & PDFormer & STID & STAEformer & ST-MambaSync\\
\hline
\hline
\multicolumn{14}{c}{METR-LA(N=325)} \\
\hline
(15 min) & MAE & 6.80 & 2.69 & 2.67 & 2.85 & 2.75 & 2.75 & 2.69 & 2.81 & 2.80 & 2.83 & 2.82 & 2.65 & \textcolor{red}{\textbf{2.63}}\\
& RMSE & 14.21 & 5.15 & 5.16 & 5.53 & 5.29 & 5.27 & 5.16 & 5.57 & 5.55 & 5.45 & 5.53 & 5.11 & \textcolor{red}{\textbf{5.05}}\\
& MAPE & 16.72 & 6.99 & 6.86 & 7.63 & 7.10 & 7.12 & 6.89 & 7.40 & 7.41 & 7.77 & 7.75 & 6.85 & \textcolor{red}{\textbf{6.80}}\\
(30 min) & MAE & 6.80 & 3.08 & 3.12 & 3.20 & 3.15 & 3.14 & 3.05 & 3.18 & 3.12 & 3.20 & 3.19 & 2.97 &\textcolor{red}{\textbf{2.91}}\\
& RMSE & 14.21 & 6.20 & 6.27 & 6.52 & 6.35 & 6.33 & 6.13 & 6.59 & 6.49 & 6.46 & 6.57 & \textcolor{red}{\textbf{6.00}} & 6.07\\
& MAPE & 16.72 & 8.47 & 8.42 & 9.00 & 8.62 & 8.62 & 8.16 & 8.47 & 8.73 & 9.19 & 9.39 & 8.13 & \textcolor{red}{\textbf{8.08}}\\
(60 min) & MAE & 6.80 & 3.51 & 3.54 & 3.59 & 3.60 & 3.59 & 3.47 & 3.57 & 3.44 & 3.62 & 3.55 & 3.34 & \textcolor{red}{\textbf{3.31}}\\
& RMSE & 14.20 & 7.28 & 7.47 & 7.45 & 7.43 & 7.44 & 7.21 & 7.51 & 7.35 & 7.47 & 7.55 & \textcolor{red}{\textbf{7.02}} & \textcolor{red}{\textbf{7.02}}\\
& MAPE & 10.15 & 9.96 & 10.32 & 10.47 & 10.35 & 10.25 & \textcolor{red}{\textbf{9.70}} & 10.24 & 10.07 & 10.91 & 10.95 & \textcolor{red}{\textbf{9.70}} & \textcolor{red}{\textbf{9.70}}\\
\hline
\multicolumn{14}{c}{PEMS-BAY(N=207)} \\
\hline
(15 min) & MAE & 3.06 & \textcolor{red}{\textbf{1.30}} & 1.31 & 1.35 & 1.36 & 1.37 & 1.33 & 1.33 & 1.35 & 1.32 & 1.31 & 1.31 & \textcolor{red}{\textbf{1.30}}\\

& RMSE & 7.05 & \textcolor{red}{\textbf{2.73}} & 2.76 & 2.88 & 2.88 & 2.92 & 2.80 & 2.82 & 2.90 & 2.83 & 2.79 & 2.78 & 2.75\\

& MAPE & 6.85 & \textcolor{red}{\textbf{2.71}} & 2.73 & 2.91 & 2.86 & 2.85 & 2.81 & 2.76 & 2.87 & 2.78 & 2.78 & 2.76 & 2.75\\

(30 min) & MAE & 3.06 & 1.63 & 1.65 & 1.67 & 1.70 & 1.72 & 1.66 & 1.65 & 1.65 & 1.64 & 1.64 & 1.62 & \textcolor{red}{\textbf{1.63}}\\

& RMSE & 7.04 & 3.73 & 3.75 & 3.82 & 3.84 & 3.86 & 3.77 & 3.77 & 3.82 & 3.79 & 3.73 & 3.68 & \textcolor{red}{\textbf{3.62}}\\

& MAPE & 6.84 & 3.73 & 3.71 & 3.81 & 3.79 & 3.88 & 3.75 & 3.66 & 3.74 & 3.71 & 3.73 & 3.62 & \textcolor{red}{\textbf{3.61}}\\

(60 min) & MAE & 3.05 & 1.99 & 1.97 & 1.94 & 2.02 & 2.06 & 1.95 & 1.92 & 1.91 & 1.91 & 1.91 & 1.88 & \textcolor{red}{\textbf{1.87}}\\

& RMSE & 7.03 & 4.60 & 4.60 & 4.50 & 4.63 & 4.60 & 4.50 & 4.45 & 4.49 & 4.43 & 4.42 &  4.34& \textcolor{red}{\textbf{4.30}}\\

& MAPE & 6.83 & 4.71 & 4.68 & 4.55 & 4.72 & 4.88 & 4.62 & 4.46 & 4.52 & 4.51 & 4.55 & 4.41 & \textcolor{red}{\textbf{4.40}}\\
\hline
\end{tabular}\label{tab:horizon}
\end{table}

\subsection{Ablation Study}
In this subsection, we conduct various ablation studies, including adjustments to the layer configurations in both the attention-based and Mamba-based models. Additionally, we analyze the trade-offs between accuracy and computational efficiency.

\subsubsection{Vanellia Ablation Study}
\begin{table}[t]
\centering
\caption{Performance comparison on the PEMS08 dataset}
\begin{tabular}{@{}lcccccc@{}}
\toprule
\textbf{Model} & \textbf{MAE} & \textbf{RMSE} & \textbf{MAPE} & \textbf{FLOPS(M)} & \textbf{Inference (s)} & \textbf{Train (s)} \\ \midrule
STAEformer (attention 3 layers) & 13.49 & 23.30 & 8.84 & 4.24 & 3.03 & 36 \\
STAEformer (attention 2 layers) & 13.54 & 23.47 & 8.887 & 2.87 & 2.09 & 23 \\
STAEformer (attention 1 layer) & 13.77 & 23.27 & 9.16 & 1.49 & 1.20 & 14 \\
\hline
ST-SSMs(mamba 3 layer)	&13.45	&23.08&	8.96	&1.07	&3.64	&42\\
ST-SSMs(mamba 2 layer)	&13.43	&23.14&	8.95	&0.75	&2.56	&28\\
ST-SSMs (mamba 1 layer) & 13.40 & 23.20 & 9.00 & 0.43 & 1.18 & 14 \\
\hline
ST-MambaSync (mamba 1 layer \& attention 1 layer) & 13.30 & 23.144 & 8.80 & 1.49 & 2.65 & 29 \\ 
ST-MambaSync (mamba 1 layer \& attention 2 layer) & 13.37 & 23.42 & 8.98 & 2.87 & 3.40 & 33 \\
ST-MambaSync (mamba 2 layer \& attention 1 layer) & 13.45 & 24.16 & 10.98 &  1.49 & 2.96 & 30 \\
\bottomrule
\end{tabular}
\label{tab:performance_comparison}
\end{table}

We conducted a comparative analysis with the attention-based model (STAEFormer) \citep{staeformer}, the Mamba-based model (ST-SSMs) \citep{shao2024stssms}, and our newly proposed hybrid model (ST-MambaSync). Notably, STAEFormer employs three attention layers. To investigate the impact of the number of attention layers on performance, we modified the STAEFormer by varying the number of layers. As illustrated in Table~\ref{tab:performance_comparison}, the comparison underscores the trade-offs between prediction accuracy and computational efficiency across the models. STAEFormer, with multiple attention layers, achieves a competitive mean absolute error (MAE) range of 13.49 to 13.77 but requires substantial computational resources (8.84 to 9.16 FLOPS) and exhibits longer inference (1.20s to 3.03s) and training durations (14s to 36s). Conversely, ST-SSMs, which incorporates a single Mamba layer and no attention layers, shows a high accuracy (MAE of 13.40) with significantly lower computational demands (FLOPS of 0.43), making it an effective option for efficient traffic flow prediction. The ST-MambaSync model, combining one Mamba layer with one attention layer, outperforms others by achieving the lowest prediction error (MAE of 13.30) while maintaining reasonable computational efficiency (1.49 FLOPS). Its inference and training times are 2.65s and 29s, respectively, indicating its suitability for practical implementation in real-world traffic management systems. Increasing the number of Mamba or attention layers in ST-MambaSync, however, results in diminished performance compared to configurations with a single layer of each type. Nevertheless, the integrated approach of combining Mamba and attention layers generally surpasses models utilizing only one of these mechanisms.

\subsubsection{Trade Off Analysis on Accuracy and Computation}
\begin{figure}[t]
    \centering   
    \includegraphics[scale = 0.4]{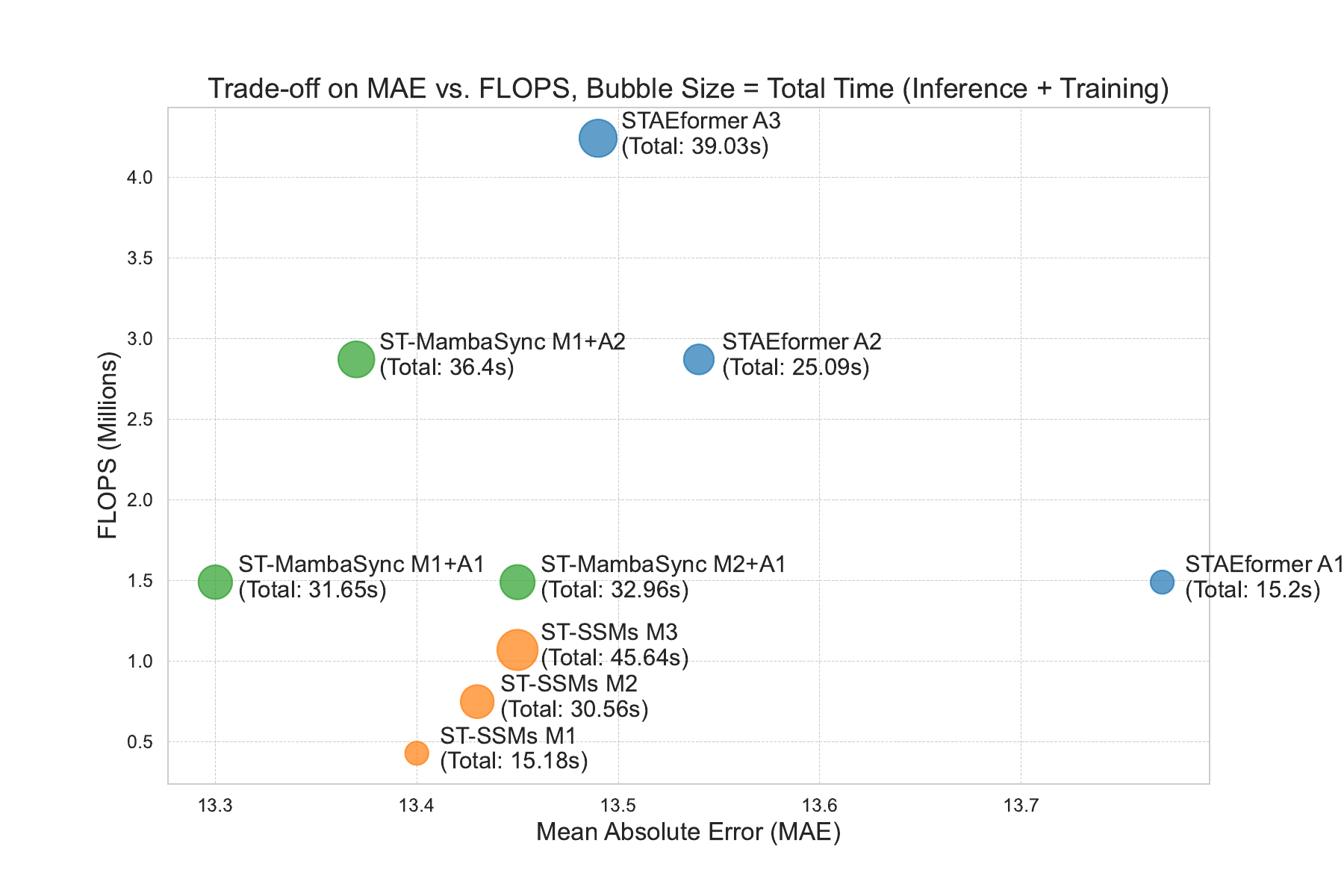}
    \caption{Trade-offs in Model Performance and Computational Efficiency. This bubble chart illustrates the relationship between Mean Absolute Error (MAE) and computational cost (FLOPS) for various predictive models on the PEMS08 dataset. Each bubble's size represents the total time required for inference and training, highlighting the efficiency trade-offs. We denote ``M'' as the number of Mamba layer in the model,  ``A'' as the number of attention layers. }
    \label{fig:tradeoff}
\end{figure}

We further conduct a trade-off analysis on the attention-based model and the Mamba-based model, which aids in investigating the combined efficiency of our proposed ST-MambaSync. The results of the trade-off analysis on accuracy and computational effectiveness are presented in Figure~\ref{fig:tradeoff}. The analysis begins by examining two aspects: the increase in attention layers and the increase in Mamba layers.

\paragraph{Increase in Attention Layers}
In attention-based models, an increase in the number of attention layers leads to higher FLOPS and computational times, although with improved accuracy. For our integrated model, ST-MambaSync, maintaining a single Mamba layer while increasing the number of attention layers results in a doubling of FLOPS and a significant increase in computational time, without a corresponding improvement in accuracy.

\paragraph{Increase in Mamba Layers}
For the Mamba-based model, ST-SSMs, an increase in Mamba layers results in higher FLOPS and extended computational times, but surprisingly, accuracy decreases. This observation suggests that a single Mamba layer is optimal for balancing accuracy and computational efficiency in the integrated ST-MambaSync model. To enhance prediction accuracy without significantly increasing computational burden, incorporating a single attention layer is the most effective strategy. The optimal configuration for achieving the highest accuracy with the least computational trade-off is the ST-MambaSync model with one Mamba layer and one attention layer, which meets these conditions satisfactorily.

\subsubsection{Temporal Analysis of Traffic Flow Predictions}

\begin{figure}[h]
    \centering
\includegraphics[scale = 0.3]{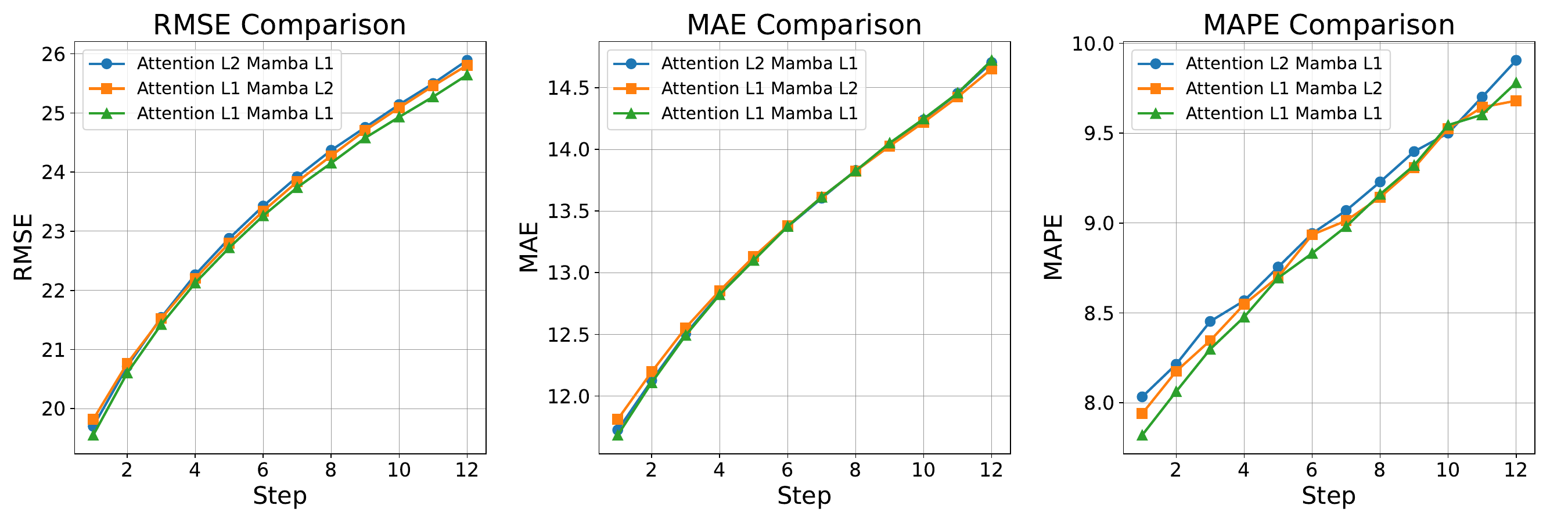}
    \caption{This figure presents a side-by-side comparison of three key performance metrics: Root Mean Square
Error (RMSE), Mean Absolute Error (MAE), and Mean Absolute Percentage Error (MAPE), across varying layers of attention and mamba for ST-MambaSync. Each subplot illustrates the
variation of a specific metric across 12 time steps, highlighting the models’ performance stability and accuracy in
forecasting. Distinct color-coded lines represent different model configurations, ensuring clear differentiation and
readability.}
    \label{fig:layers_compare}
\end{figure}

In Figure \ref{fig:layers_compare}, we present an hour-long forecast for PEMS08, visualizing predictions for each
5-minute time step for the metrics RMSE, MAE, and MAPE.The figure presents a comparison of performance metrics for ST-MambaSync: "Attention L2 Mamba L1" and "Attention L1 Mamba L2". The metrics evaluated include Root Mean Square Error (RMSE), Mean Absolute Error (MAE), and Mean Absolute Percentage Error (MAPE) across different prediction steps. In all three metrics, both models exhibit similar trends, with RMSE and MAE increasing gradually with each step, while MAPE shows slight fluctuations. Notably, "Attention L2 Mamba L1" consistently outperforms "Attention L1 Mamba L2" across all metrics and steps, demonstrating its superiority in predicting traffic flow dynamics. The comparison underscores the effectiveness of leveraging both attention and Mamba blocks in enhancing prediction accuracy. Furthermore, it highlights the importance of model architecture in achieving superior performance in traffic forecasting tasks. These findings contribute to the advancement of traffic prediction methodologies, with implications for real-world applications in urban planning and traffic management systems.

\section{Discussion and Implication}\label{Discussion}

\begin{figure}[H]
    \centering
\includegraphics[scale = 0.3]{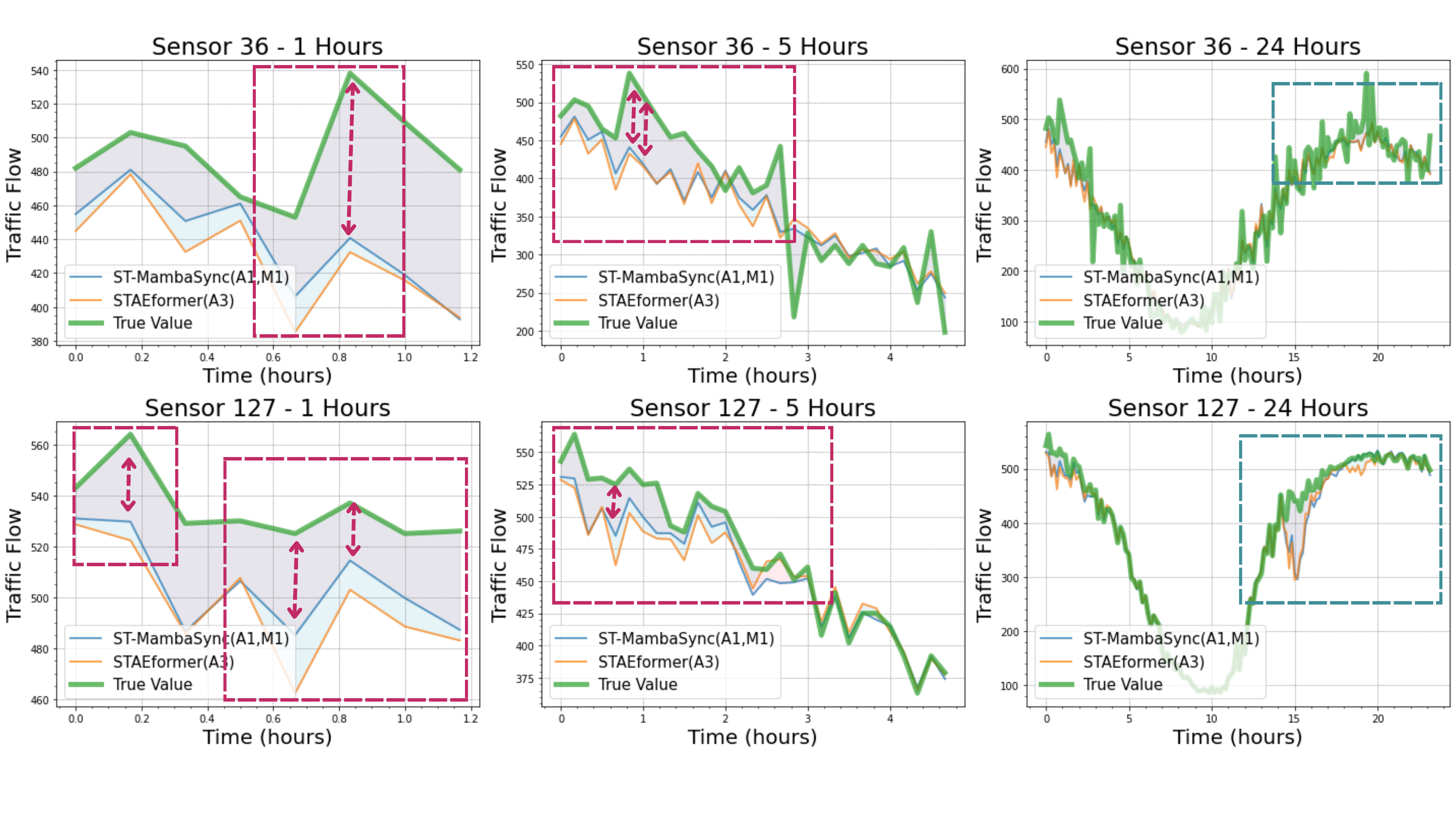}
    \caption{Comparative Analysis of Prediction Results Using PEMS08 Dataset for Sensors 36 and 127.}
    \label{fig:case}
\end{figure}

This case study evaluates the predictive performance of the STAEformer and ST-MambaSync models across 1-hour, 5-hour, and 24-hour intervals. Each model's architecture is defined by the number of attention layers (denoted as "A\#") and the number of Mamba layers ("M\#"). For instance, ST-MambaSync is configured with one attention layer and one Mamba layer, while STAEformer utilizes three attention layers, as illustrated in Figure~\ref{fig:case}.

\paragraph{Predictions for Sensor 36}
\begin{itemize}
    \item \textbf{1 Hour:} Both models diverged from the true values, though ST-MambaSync demonstrated closer approximations at certain intervals.
    \item \textbf{5 Hours:} ST-MambaSync provided more consistent and accurate predictions, closely tracking the actual data, unlike the fluctuating results from STAEformer.
    \item \textbf{24 Hours:} ST-MambaSync showed superior long-term predictive consistency, adhering closely to the actual traffic flow patterns.
\end{itemize}

\paragraph{Predictions for Sensor 127}
\begin{itemize}
    \item \textbf{1 Hour:} Both models performed similarly to those for Sensor 36, with ST-MambaSync slightly more accurate at certain points.
    \item \textbf{5 Hours:} Variability was noted in both models, with neither showing a consistent advantage during the initial hours.
    \item \textbf{24 Hours:} ST-MambaSync maintained closer alignment with the true values, indicating its better capability at handling longer-term dynamics.
\end{itemize}
The analysis indicates that ST-MambaSync tends to provide more accurate and consistent forecasts across all examined intervals for both sensors. It excels particularly in the 24-hour forecasts, suggesting it is more adept at capturing and adapting to longer-term traffic flow dynamics. ST-MambaSync consistently outperforms the state of art model (SOTA) STAEformer, especially in longer forecast intervals. This emphasizes the importance of selecting appropriate models based on the predictive timeframe and desired accuracy level for traffic management applications.

\section{Conclusion}\label{Conclusion}
In summary, this study introduces ST-MambaSync, a novel framework blending an optimized attention layer with a simplified state-space layer. Our experiments show that ST-MambaSync achieves state-of-the-art accuracy in spatial-temporal prediction tasks while minimizing computational costs. By revealing Mamba's function akin to attention within a residual network structure, we highlight the efficiency of state-space models. This balance between accuracy and efficiency holds promise for various real-world applications, from urban planning to traffic management. 


\bibliographystyle{unsrtnat}
\bibliography{example}

\end{document}